\documentclass{article}

\usepackage{PRIMEarxiv}

\usepackage[utf8]{inputenc} 
\usepackage[T1]{fontenc}    
\usepackage{hyperref}       
\usepackage{url}            
\usepackage{booktabs}       
\usepackage{amsfonts}       
\usepackage{nicefrac}       
\usepackage{microtype}      
\usepackage{lipsum}
\usepackage{fancyhdr}       
\usepackage{graphicx}       
\usepackage{amsmath} 
\graphicspath{{media/}}     

\title{Simplifying CLIP: Unleashing the Power of Large-Scale Models on Consumer-level Computers
}

\author{
  Hongbo Liu \\
  Tongji University\\
  \texttt{1952350@tongji.edu.cn}
 \And
  Shengjie Zhao  \\
  Tongji University \\
  \texttt{shengjiezhao@tongji.edu.cn}\\
  \AND
  Weichao Chen \\
  Tongji University \\
  \texttt{carlyle.chen@tongji.edu.cn} \\
  \And
  Deniz Gunduz \\
  Imperial College London \\
  \texttt{d.gunduz@imperial.ac.uk} \\
}

\begin{document}
\maketitle

\begin{abstract}
Contrastive Language-Image Pre-training (CLIP) has attracted a surge of attention for its superior zero-shot performance and excellent transferability to downstream tasks. However, training such large-scale models usually requires substantial computation and storage, which poses barriers for general users with consumer-level computers. Motivated by this observation, in this paper we investigate how to achieve competitive performance on only one Nvidia RTX3090 GPU and with one terabyte for storing dataset. On one hand, we simplify the transformer block structure and combine Weight Inheritance with multi-stage Knowledge Distillation (WIKD), thereby reducing the parameters and improving the inference speed during training along with deployment. On the other hand, confronted with the convergence challenge posed by small dataset, we generate synthetic captions for each sample as data augmentation, and devise a novel Pair Matching (PM) loss to fully exploit the distinguishment among positive and negative image-text pairs. Extensive experiments demonstrate that our model can achieve a new state-of-the-art datascale-parameter-accuracy tradeoff, which could further popularize the CLIP model in the related research community.
\end{abstract}

%

\section{Introduction}

Pre-trained large image-text foundation models, famous as the contrastive language-image pre-training (CLIP) model \cite{radford2021learning}, have recently drawn considerable attention in the fields of computer vision and natural language processing. These models have demonstrated excellent zero-shot performance and robustness across a wide range of downstream tasks, such as image-text retrieval and classification (Zhu et al. 2023). However, the substantial computational and storage costs for training CLIP-like models impeded their further popularization. For instance, MobileCLIP \cite{vasu2024mobileclip} was trained on 256×A100 GPUs with a global batch size of 65,536, and the corresponding dataset DataCompDR-1B requires 140 TB local storage space. In addition, the huge parameter size (e.g., CLIP-B/16 model \cite{radford2021learning} consists of 86.2M parameters for the image encoder and 63.4M parameters for the text encoder) contributes to increasing inference latency, which raises challenges for deploying the model on devices with limited computational resources. These shortcomings pose barriers for general users with insufficient computational resources and datasets to engage in the training and deploying of such large-scale models.


In practice, the GPU memory of consumer-level computers typically does not exceed 24GB (e.g., Nvidia RTX3090), while storage capacity could be less than 1TB. In the context of training CLIP-like models under such resource constraints, two primary problems have to be addressed. Firstly, it is imperative to minimize the number of parameters that need training, while retaining existing model knowledge as much as possible. Secondly, small-scale datasets need to be augmented appropriately, and more effective methods need to be developed to fully exploit the internal correlation of image-text pairs within the limited samples.

\begin{figure}[t]
\centering
\includegraphics[width=1.0\linewidth]{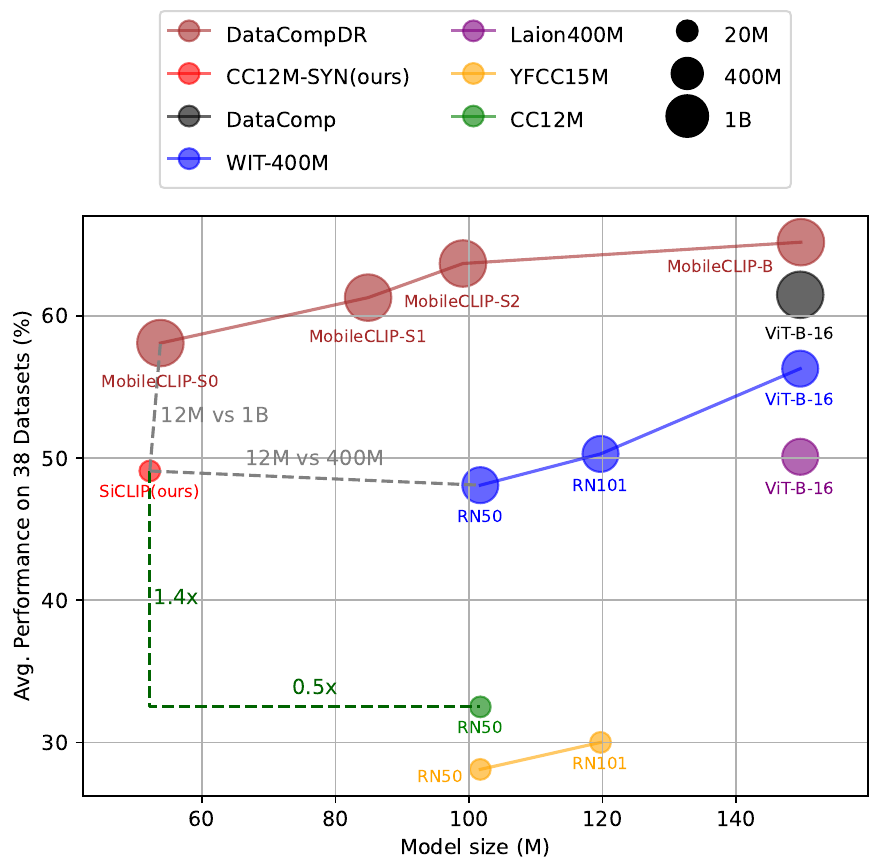}
\caption{Average zero-shot performance on downstream tasks. Compared to current works, our model achieves competitive performance with a smaller model size while being trained on a small-scale (12M) dataset.}
\label{fig:performance_comparison}
\end{figure}

In this paper, we investigate how to train a lightweight CLIP model using only one RTX3090 GPU and 1TB of storage, thereby popularizing the research of CLIP-like models on consumer-level computers. To this end, we first propose to simplify the traditional transformer block as SAS-P block, together with applying the strategy of weight sharing. Then, by inheriting weights and distilling knowledge from existing models, the number of parameters required for training can be reduced further. In terms of the dataset, we choose the widely used CC12M \cite{changpinyo2021conceptual} as a basis. The dataset not only suffers from its small scale but also exhibits low-quality of its labels, which both bring difficulty on convergence of model training process. To address this issue, we enhance each image samples of the CC12M with multiple text labels, creating the new CC12M-SYN. Furthermore, to extract valuable information from such small dataset, we introduce the Pair Matching (PM) loss to help the model capture the distinguishment among positive and negative image-text pairs. These methods significantly improve the convergence speed of model training with our extensive experiments. Finally, through performance comparisons across 38 datasets with several popular approaches (Fig. \ref{fig:performance_comparison}), our proposed SiCLIP framework achieves a new state-of-the-art datascale-parameter-accuracy trade-off.

\paragraph{Our contributions:}
The contributions of this work can be summarized as follows:
\begin{itemize}
    \item We propose a systematic framework for training lightweight CLIP models on a consumer-level computer, including the dataset construction and the corresponding training procedure, named as SiCLIP. In SiCLIP, both the computational and storage cost have been reduced, while retaining competitive performance compared to other large-scale models. 
    \item We simplify the CLIP model structure by sharing weights among SAS-P blocks, and combine Weight Inheritance with multi-stage Knowledge Distillation (WIKD), thereby reducing the memory requirements when training and deploying.
    \item  A new loss function called PM loss is devised, which predicts whether an image-text pair is matched or not during training. Coupling with our augmented dataset CC12M-SYN, the PM loss can exploit the distinguishment among positive and negative image-text pairs. Experimental results show that both the new dataset and PM loss can improve the training efficiency significantly while increasing the dataset size marginally.
\end{itemize}

\begin{figure*}[htbp]
\centering
\includegraphics[width=15cm]{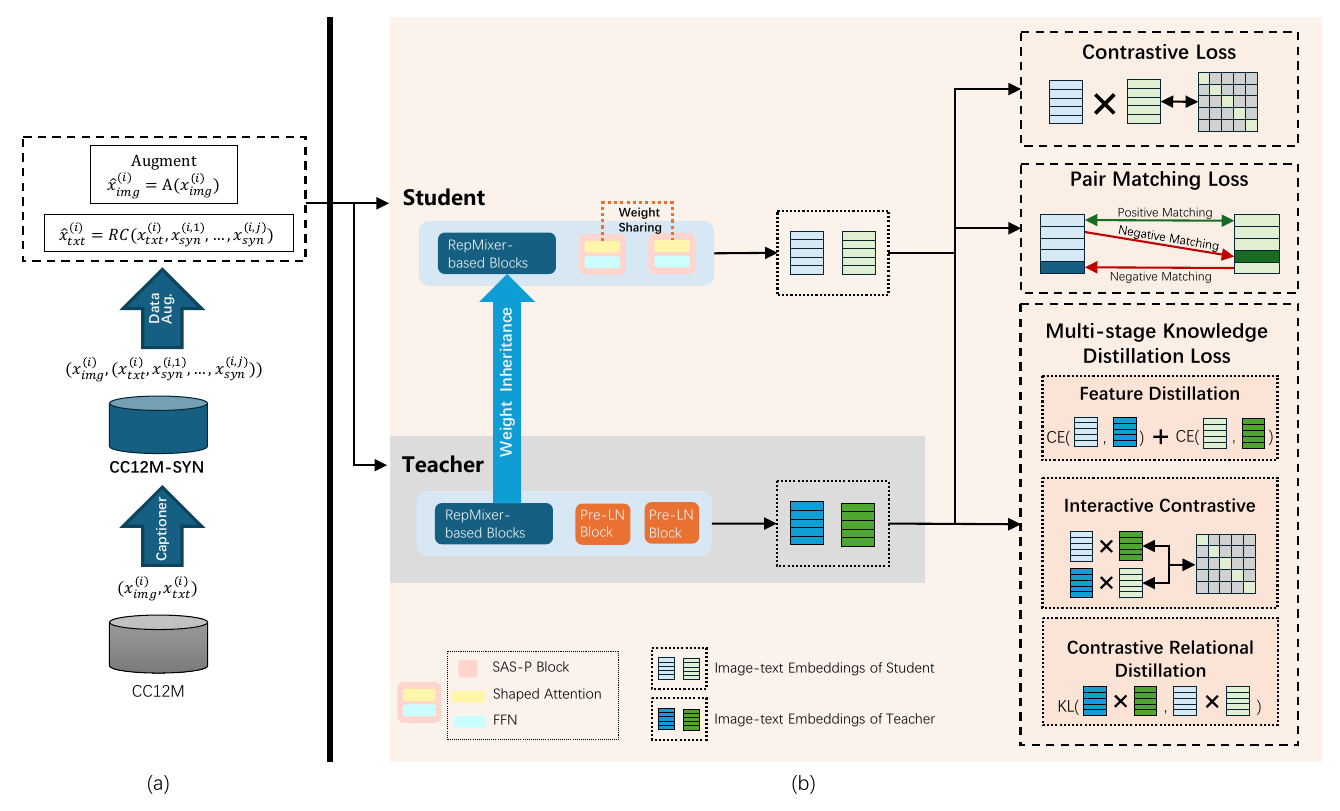}
\caption{Overview of our proposed framework SiCLIP. (a) We augment the widely used CC12M dataset by adding multiple synthetic captions for each image. During training, for each image-text pair, the image is augmented by RandomResizedCrop and RandAugment, and the text is randomly chosen (RC) from the set of original and synthetic captions. (b) Illustration of training procedure. We simplify the model structure, inherit the weights from the teacher, and optimize contrastive loss, PM loss and multi-stage KD loss simultaneously during training.}
\label{fig:SiCLIP}
\end{figure*}

\section{Related Work}

\subsection{Efficient Training for CLIP}
Ever since the introduction of CLIP as a large image-text foundation model with impressive zero-shot performance for a wide variety of downstream tasks, there have been a number works aiming to improve its training efficiency and model size. Examples include fine-grained image-text alignment \cite{yao2021filip}, data augmentation \cite{mu2022slip,li2021supervision,vasu2024mobileclip}, unimodal self-supervision \cite{mu2022slip,li2021supervision}, and contrastive learning in image-text-label space \cite{yang2022unified}. In addition, Zhai et al. \cite{zhai2023sigmoid} propose the pairwise sigmoid loss as a simple alternative to contrastive loss, demonstrating its effectiveness when training with a small batch size. However, it may cause quadratic computational complexity because the matching logits are calculated between all positive and negative image-text pairs. Li et al. \cite{li2022blip} use fine-grained image-text matching (ITM) loss as a complement to the contrastive loss, but ITM needs a multi-layer transformer based encoder to encode the multimodal fine-grained features, which is not suitable for lightweight models. 


Weight Inheritance (WI) and Knowledge Distillation (KD) \cite{hinton2015distilling} based methods are also adopted for efficient training. TinyCLIP \cite{wu2023tinyclip} trains compact CLIP models via cross-modal affinity mimicking and WI. Yang et al. \cite{yang2024clip} explore the effectiveness of different KD methods for CLIP training. 

Quality of datasets is also important for efficient training. Fang et al. \cite{fang2023data} and Gadre et al. \cite{gadre2024datacomp} have exploited filtering methods to remove noisy samples. However, remaining captions may still not be descriptive enough. Recent works \cite{yang2023alip,lai2023scarcity} show that synthetic captions generated from a pre-trained captioning model can improve dataset quality.  

\subsection{Simplifying the Transformer Architecture}

Following the remarkable success of transformers in a large variety of tasks, recent years have witnessed many efforts to simplify the transformer architecture to increase their training and inference efficiency. Yu et al. \cite{yu2022metaformer} demonstrated that the general structure of transformer blocks is more essential for the performance, making it possible to eliminate the attention-based token mixers, which are usually prohibitively expensive due to the quadratic complexity of multi-head self-attention (MHSA) over a long sequence of representations. Besides, previous studies in both CNNs and transformers have shown that shallow layers mainly focus on local patterns while deeper layers tend to capture high-level semantics or global relationships \cite{hou2017deeply, wu2019cascaded, vitdosovitskiy2020image}, thus it is usually unnecessary to model global relationship by MHSA at the early stages. Based on these facts, Liu et al. \cite{liu2021swin} proposed a hierarchical transformer, and employed shifted windows to limit self-attention computation to non-overlapping local windows while also allowing for cross-window connection, bringing greater efficiency. In a different line of work, Pan et al. \cite{pan2022less} and Guo et al. \cite{guo2022cmt} introduce convolutional layers into early transformer layers. Following these works, Vasu et al. \cite{vasu2023fastvit} proposed RepMixer as a token mixer, which uses structural reparameterization to lower the memory access cost by removing skip-connections in the network. 

As a simple but effective lightweight method, weight-sharing strategy has been employed in many transformer-based models. Dehghani et al. \cite{dehghani2018universal} first proposed the idea of reusing the transformer layer for natural language processing tasks with a different motivation: they regard repeated network layers as a complementary way of introducing a recurrent inductive bias to transformers, and it was observed that their method outperforms the vanilla transformer on several tasks. Jaegle et al. \cite{jaegle2021perceiver} adopted the cross-attention layer weight sharing in multimodal pre-training. Hernandez et al. \cite{hernandez2023sharing} explored sharing different parts of conformer \cite{gulati2020conformer} at different levels of granularity with the model size hard-constrained by memory. Most recently, He et al. \cite{he2024simplifying} studied the standard Pre-LN transformer block \cite{vaswani2017attention} through signal propagation theory and proposed a simplified parallel structure transformer block equipped with shaped attention \cite{noci2024shaped} as token mixers, named as Simplified Attention Sub-block Parallel (SAS-P), reducing the number of parameters and increasing throughput of the model without performance loss on language downstream tasks. Our work is the first attempt on extending SAS-P into the multi-modal domain, and further simplify it by sharing weights of the token mixers between adjacent blocks.

\section{Methods}
In this section, we first introduce our simplified model structure, which shares weights among SAS-P blocks. Then, we introduce a method called WIKD for efficient training. Next, we introduce a novel loss function, called the pair matching (PM) loss, to further improve the training performance. Finally, we also improve the CC12M dataset, which we use for training our model, by adding synthetic captions to improve data diversity and data quality, with minimal additional storage. The new dataset is called CC12M-SYN. Fig. \ref{fig:SiCLIP} shows the overall framework of our method.

\subsection{Simplifying Model Structure by Sharing Weights Among SAS-P Blocks}
\begin{figure}[t]
\centering
\includegraphics[width=0.8\linewidth]{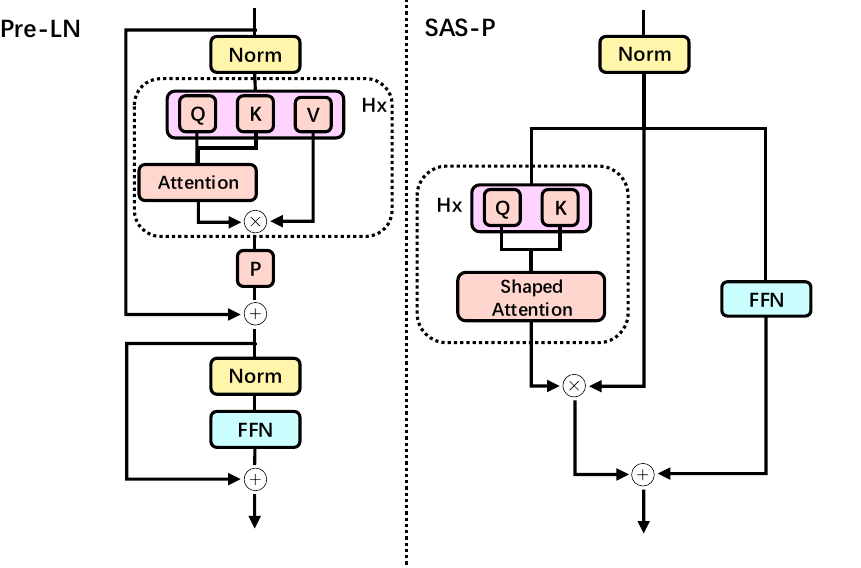}
\caption{Pre-LN block \cite{vaswani2017attention} vs SAS-P block \cite{he2024simplifying}. SAS-P block removes skip connections, value and projection parameters, leading to faster inference speed with fewer parameters than Pre-LN.}
\label{fig:preln_vs_sasp}
\end{figure}

Our architecture is built upon the state-of-the-art MobileCLIP-S0 model \cite{vasu2024mobileclip}, which we enhance in multiple ways. The MobileCLIP-S0 framework features hybrid structures for both the image encoder and text encoder, incorporating a synergistic arrangement of convolution-based and MHSA-based blocks. However, for each MHSA-based block, MobileCLIP-S0 simply adopts the standard Pre-LN block with MHSA as the token mixer \cite{vaswani2017attention}, as shown in Fig. \ref{fig:preln_vs_sasp} (left). 

We start by reducing the parameters of the skip connections within each Pre-LN block. The presence of these connections causes a bottleneck on memory access and inference speed, and a lightweight MHSA-based block design becomes essential. Moreover, it has been shown that the feed-forward layers can be integrated into the attention module without degrading performance of the transformer layer \cite{sukhbaatar2019augmenting, niu2024beyond}.


The right-hand side of Fig. \ref{fig:preln_vs_sasp} illustrates SAS-P (He and Hofmann 2024), a simplified parallel transformer block that eliminates skip connection along with the value and projection parameters. It uses shaped attention \cite{noci2024shaped} as its token mixer to prevent signal degradation after removing skip connections, making attention matrices more identity-like to maintain a good signal propagation. The attention matrices of shaped attention are given by: 
\begin{gather}
    A(X) = \text{Softmax}\left(\frac{1}{\sqrt{d_v}} XW^QW^{K^T}X^T\right), \label{eq:sas_a}\\
    A(X) \leftarrow \alpha I_T + \beta A(X) -\gamma C,
\end{gather}
where $X$ denotes the SAS-P input, $W^Q$ and $W^K$ are the query and the key matrices, $d_v$ is the model dimension, $I_T$ is the identity matrix, and $\alpha$, $\beta$, $\gamma$ are learnable parameters. $C$ is the center matrix, and each element of $C$ is set to $\frac{1}{n}$, with $n$ being the number of input tokens. At initialization, $W^Q$ is set to $0$, $\alpha$, $\beta$, $\gamma$ are set to $1$, leading to $\beta A(X)-\gamma C = 0$ and $A(X) = I_T$, which is effective for good signal propagation. SAS-P shows impressive performance on several language tasks \cite{he2024simplifying} while achieving faster inference speed with fewer parameters than Pre-LN.

\begin{figure}[htbp]
\centering
\includegraphics[width=0.7\linewidth]{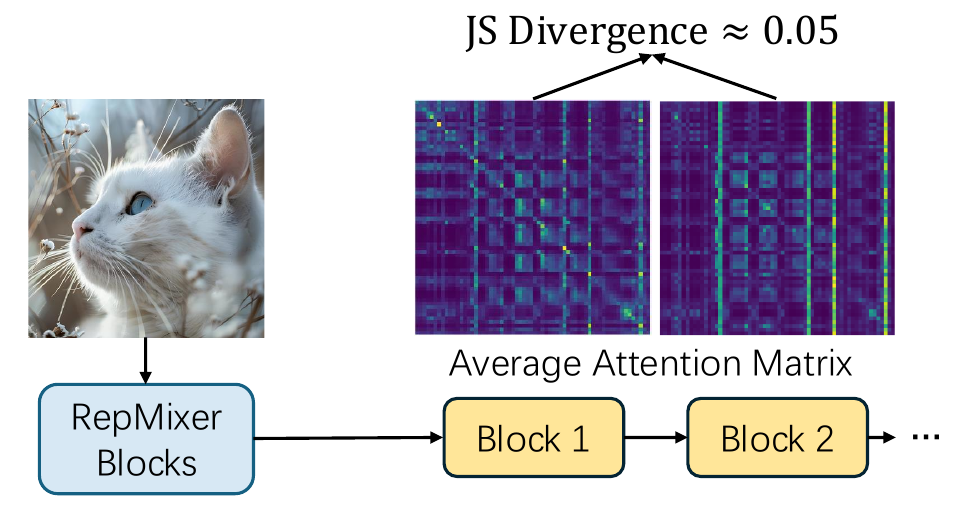}
\caption{Illustration of JS divergence of the average multi-head attention matrices between adjacent blocks. 
}
\label{fig:avg_att_mat}
\end{figure}

To simplify the model structure further, we evaluate the Jensen-Shannon (JS) divergence between adjacent MHSA-based blocks (see Fig. \ref{fig:avg_att_mat}). A low JS divergence implies that weight sharing can be used among these matrices without degrading the performance.
Accordingly, before employing KD during training, our ``student'' model replaces all of the Pre-LN blocks with SAS-P blocks, and apply weight sharing among these blocks. As a result, the image encoder of our model has approximately $14\%$ fewer parameters compared to MobileCLIP-S0, and only $11\%$ of that in OpenAI-B/16 \cite{radford2021learning}.

\subsection{Weight Inheritance with Multi-Stage Knowledge Distillation (WIKD)}

To benefit from small-scale datasets, a widely used paradigm is to employ a task-related pre-trained backbone and add a few task-specific layers \cite{hu2021lora}. 
Inspired by the idea of using backbones, we adopt WI \cite{wu2023tinyclip} to train CLIP on small-scale datasets
. In practice, since we modify the MHSA-based blocks of MobileCLIP-S0 structure and keep the RepMixer-based blocks unmodified (which are already efficient), we can directly inherit the weights of these modules from MobileCLIP-S0 which has been well pretrained on a large-scale dataset \cite{vasu2024mobileclip}. In this case, the inherited modules can be seen as the `backbone'. Then, we freeze these inherited layers and only train the newly added SAS-P blocks on a much smaller dataset. Applying the above methods can reduce gradient storage, and thus allow us to use a larger batch size to maintain the performance of contrastive learning.

Moreover, we regard our model as a student model of MobileCLIP-S0, and perform multi-stage KD during training, leading to further performance improvement. Specifically, we apply KD on the unimodal feature space (stage 1), contrastive relation space (stage 2), and interactive contrastive space (stage 3). Given a batch of image-text pairs, the student model firstly mimics the image and text feature distributions of the teacher by optimizing feature distillation loss ($L_{\text{FD}}$), given by:
\begin{equation}
    L_\text{FD} = \frac{1}{|B|}\sum_{k=1}^{|B|}(||v_k^\text{T}-v_k^\text{S}||_2^2 + ||t_k^\text{T}-t_k^\text{S}||_2^2)/2 \label{eq:fd},
\end{equation}
where $(v_k^\text{T},t_k^\text{T})$ and $(v_k^\text{S},t_k^\text{S})$ represent (image, text) features of the teacher and student models, respectively, where $B$ is the batch size. 
Then, it computes the contrastive relational distillation loss ($L_{\text{CRD}}$) and interactive contrastive loss ($L_{\text{IC}}$) to mimic the image-text similarity matrix distributions in contrastive relation space and interactive contrastive space, defined as: 
\begin{gather}
    L_{\text{IC}_{I\xrightarrow{}T}} = -\sum_{k=1}^{|B|}\log\frac{\text{exp}(v_k^\text{S}\cdot t_k^\text{T}/\tau)}{\sum_{b=1,b\ne k}^{|B|}\text{exp}(v_k^\text{S}\cdot t_b^\text{T}/\tau)}, \\
    L_{\text{IC}_{T\xrightarrow{}I}} = -\sum_{k=1}^{|B|}\log\frac{\text{exp}(t_k^\text{S}\cdot v_k^\text{T}/\tau)}{\sum_{b=1,b\ne k}^{|B|}\text{exp}(t_k^\text{S}\cdot v_b^\text{T}/\tau)} \label{eq:ic},\\
    L_{\text{IC}} = \frac{1}{2}(L_{\text{IC}_{T\xrightarrow{}I}} + L_{\text{IC}_{I\xrightarrow{}T}}),
\end{gather}
\begin{equation}
    L_{\text{CRD}} = \frac{1}{|B|}\text{KL}(\text{Sim}_\text{T}||\text{Sim}_\text{S})\label{eq:crd},
\end{equation}
where $\tau$ is a learnable temperature parameter, $\text{Sim}$ refers to the similarity matrix between image and text features. Our final distillation loss is defined as:
\begin{equation}
    L_{\text{KD}} = \lambda_1 L_{\text{FD}} + \lambda_2 L_{\text{IC}} + \lambda_3 L_{\text{CRD}}, \label{eq:kd}
\end{equation}
where $\lambda_1$, $\lambda_2$, $\lambda_3$ are hyper-parameters.
 


\subsection{Pair Matching (PM) Loss}

\begin{figure}[htbp]
\centering
\includegraphics[width=1.0\linewidth]{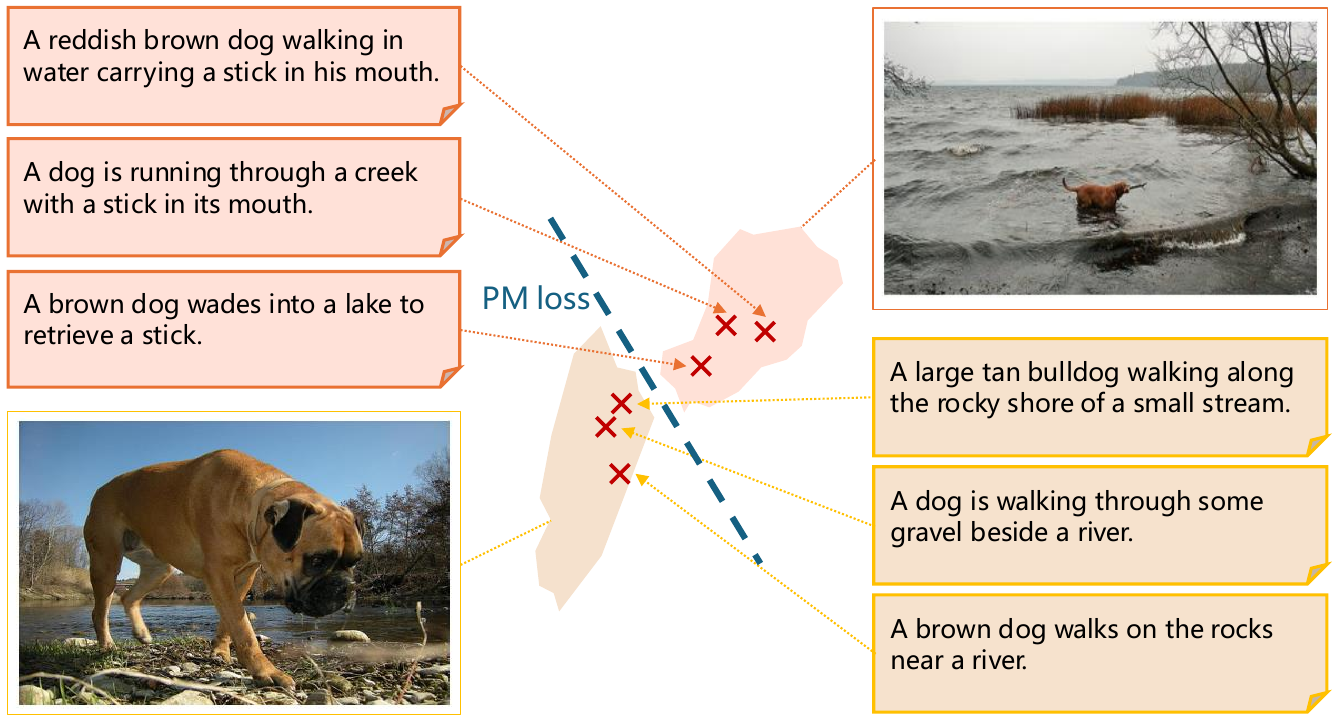}
\caption{PM loss is proposed to construct an auxiliary hyperplane (blue dashed line) to help the model distinguish semantically similar image-text pairs during training. }
\label{fig:feature_space_exam}
\end{figure}

CLIP models trained on small-scale datasets usually have unsatisfactory zero-shot performance \cite{yang2024clip}. We argue that one possible reason for this phenomenon is that models trained with fewer data have more difficulties to distinguish image-text pairs that are semantically similar. As a result, we propose to construct an auxiliary hyperplane to help the model determine whether an image-text pair is matched or not. An illustration is shown in Fig. \ref{fig:feature_space_exam}.

In particular, we add an extra binary matching task. Given a batch of image-text pairs, we firstly extract the positive image-text pairs and compute their matching logits $\hat{y}^{k(\text{pos)}}_{I\rightarrow T}$ and $\hat{y}^{k(\text{pos})}_{T\rightarrow I}$, $k=1, \ldots, B$, given by:
\begin{gather}
\hat{y}^{k(\text{pos})}_{I\rightarrow T}=\text{Linear}(v_k\cdot t_k), \\
\hat{y}^{k(\text{pos})}_{T\rightarrow I}=\text{Linear}(t_k\cdot v_k).
\end{gather}
As for the negative pairs, for each image, we select one negative text according to the image-to-text similarity matrix (one negative text has a high probability to be selected if its embedding is similar to the corresponding image). And the same process is applied to each text. As a result, the negative matching logits are defined as:
\begin{gather}
\hat{y}^{k(\text{neg})}_{I\rightarrow T}=\text{Linear}(v_k\cdot t^{(\text{neg})}), \\
\hat{y}^{k(\text{neg})}_{T\rightarrow I}=\text{Linear}(t_k\cdot v^{(\text{neg})}).
\end{gather}
Thus, our proposed PM loss is given by:
\begin{gather}
    L_{\text{PM}}=\lambda_4(\text{CE}((\hat{y}^{k(\text{pos})}_{I\rightarrow T},\hat{y}^{k(\text{neg})}_{I\rightarrow T}),\text{label})\\
    \hspace{.5in} +\text{CE}((\hat{y}^{k(\text{pos})}_{T\rightarrow I},\hat{y}^{k(\text{neg})}_{T\rightarrow I}),\text{label})),\notag
\end{gather}
and the total loss($L$) is defined as:
\begin{gather}
    L=L_{\text{CLIP}}+L_{\text{KD}}+L_{\text{PM}}.
\end{gather}

\begin{figure}[t]
\centering
\includegraphics[width=0.9\linewidth]{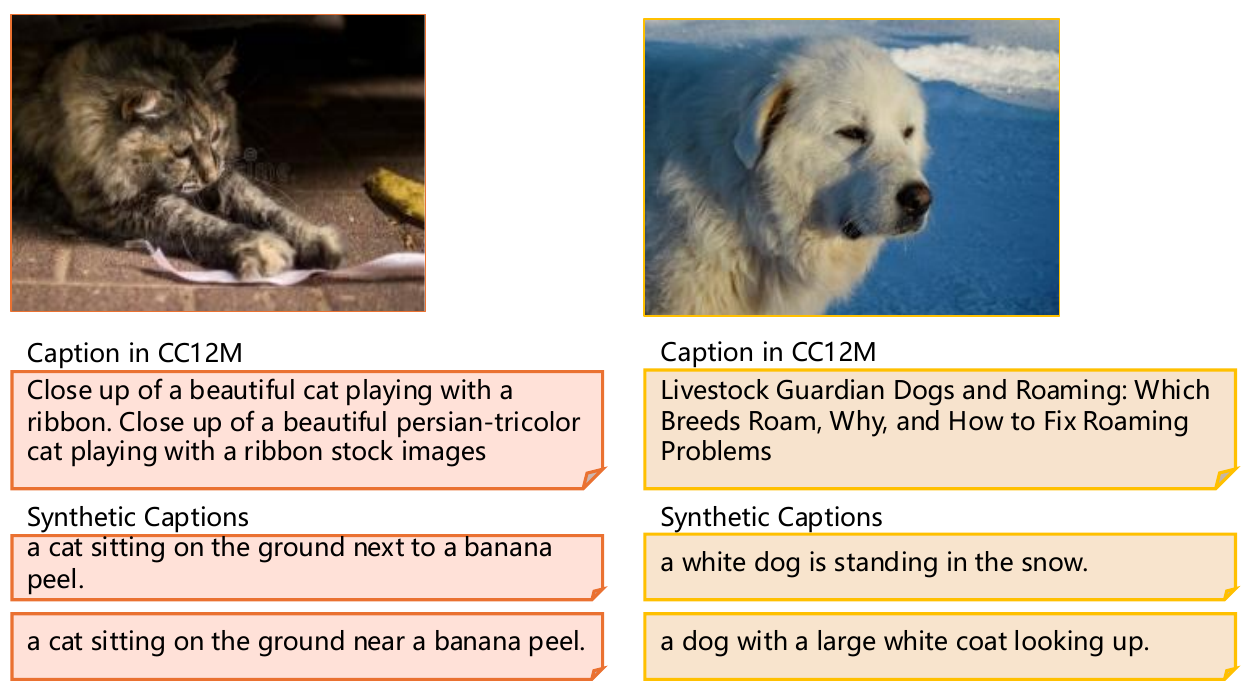}
\caption{Examples of CC12M-SYN, each sample has one original caption from CC12M and multiple synthetic captions. The synthetic captions are used to provide more descriptions of images, improving both quality and diversity of the dataset.}
\label{fig:examples_cc12msyn}
\end{figure}

\subsection{CC12M-SYN Dataset}

Image-text datasets used to train CLIP models are mostly collected from Internet, containing inherently noisy samples, which are not descriptive enough. When using small-scale datasets, the diversity and quality of data samples becomes even more important. Adding synthetic captions is a cheap but effective way to improve both diversity and quality. We adopt the widely used dataset CC12M \cite{changpinyo2021conceptual} and generate multiple synthetic captions for each image in the dataset using coca \cite{yu2022coca}, obtaining CC12M-SYN. Fig. \ref{fig:examples_cc12msyn} shows some examples with synthetic captions in CC12M-SYN. During training, we randomly select one text from the set of original and synthetic captions. As a result, one sample from CC12M-SYN consists of an image and a synthetic or original caption.

\begin{table*}[t]
  \centering

  \setlength{\tabcolsep}{2pt} 
  \renewcommand{\arraystretch}{1} 
  \begin{tabular}{@{}lcccccccc@{}}
    \toprule
    Name & {Dataset} & {Seen} & \multicolumn{2}{c}{Params (M)} & \multicolumn{2}{c}{MSCOCO} & \multicolumn{2}{c}{Flickr30k} \\
    \cmidrule(lr){4-5}\cmidrule(lr){6-7}\cmidrule(lr){8-9}
    & & Samples(B)& {Img} & {Txt} & I2T& T2I& I2T& T2I\\
    \midrule
    DataComp-B/16\cite{gadre2024datacomp}&DataComp-1B&13&86.2&63.4&\textbf{59.4}&\textbf{42.3}&\textbf{86.3}&\textbf{69.8}\\
    DataComp-B/32\cite{gadre2024datacomp}& & &86.2&63.4&53.5&37.1&79.0&61.1\\
    \midrule
    MobileCLIP-S0\cite{vasu2024mobileclip}&DataCompDR-1B&13&11.4&42.4&58.7&40.4&85.9&67.7\\

    \midrule
    TinyCLIP-63M/32\cite{wu2023tinyclip}&LAION-400M&15.8&86.2&63.4&55.5&37.6&83.2&64.4\\
    LAION-B/32\cite{schuhmann2022laion}& &(-)&86.2&63.4&53.3&35.4&79.3&62.0\\
    \midrule
    OpenAI-B/16\cite{radford2021learning}& & &86.2&63.4&58.7&40.4&85.9&67.7\\
    OpenAI-B/32\cite{radford2021learning}&WIT-400M&13&86.2&63.4&50.1&30.4&78.9&58.8\\
    OpenAI-RN50\cite{radford2021learning}& & &38.3&63.4&48.8&28.5&80.0&57.4\\
    \midrule
    \midrule
    RILS-B/16\cite{yang2023rils}&LAION-20M&0.5&86.2&37.8&32.2&25.5&45.1&34.9\\
    MaskCLIP-B/16\cite{zhou2022extract(maskclip)}& & &86.2&37.8&38.5&24.8&64.9&48.1\\
    
    \midrule
    SLIP-B/16\cite{mu2022slip}& & &86.2&37.8&31.1&20.3&57.6&40.1\\
    TinyCLIP-B/16\cite{wu2023tinyclip}&YFCC-15M&0.38&86.2&37.8&26.5&17.1&51.6&32.2\\
    TinyCLIP-39M/16\cite{wu2023tinyclip}& & &86.2&37.8&54.9&38.9&\textbf{84.4}&\textbf{66.7}\\

    \midrule
    CLIPKD-B/16\cite{yang2024clip} & & &86.2&37.8&25.0&24.7&54.6&56.6\\
    CLIPKD-RN101\cite{yang2024clip}& & &56.3&37.8&25.2&25.7&57.0&55.5\\
    CLIPKD-Swin/T\cite{yang2024clip}& CC12M+CC3M &0.48&27.9&21.3&28.5&28.6&62.2&60.9\\
    CLIPKD-MobileNetV3\cite{yang2024clip}& & &2.0&21.3&17.9&16.0&42.4&42.3\\
    CLIPKD-RN18\cite{yang2024clip}& & &11.4&21.3&21.3&19.8&47.8&47.1\\
    \midrule
    
    \textbf{SiCLIP(ours)}&CC12M-SYN&\textbf{0.38}& \underline{9.78}&\underline{42.4}&\textbf{\underline{55.7}}&\textbf{\underline{39.7}}&\underline{82.0}&\underline{66.6}\\
    
    \bottomrule
  \end{tabular}
    \caption{Zero-shot retrieval performance compared with existing models.}
      \label{tab:retrieval}
\end{table*}

\begin{table*}[t]
  \centering
  \setlength{\tabcolsep}{2pt} 
  \renewcommand{\arraystretch}{1} 
  \begin{tabular}{@{}lcccccccc@{}}
    \toprule
    Name & {Dataset} & {Seen} & \multicolumn{2}{c}{Params (M)} & IN-1k& INV2 & IN-R & IN-S \\
    \cmidrule(lr){4-5}
    & & Samples(B)& {Img} & {Txt} & & & & \\

        \midrule
    DataComp-B/16\cite{gadre2024datacomp}&DataComp-1B&13&86.2&63.4&\textbf{73.5}&\textbf{66.0}&\textbf{83.6}&\textbf{60.4}\\
    DataComp-B/32\cite{gadre2024datacomp}& & &86.2&63.4&69.2&60.8&78.2&56.8\\
    \midrule
    MobileCLIP-S0\cite{vasu2024mobileclip}&DataCompDR-1B&13&11.4&42.4&67.8&59.9&78.6&55.5\\

        \midrule

    TinyCLIP-63M/32\cite{wu2023tinyclip}& &15.8&86.2&63.4&63.9&55.7&74.1&50.8\\
    LAION-B/16\cite{schuhmann2022laion}&LAION-400M&(-)&86.2&63.4&67.0&59.4&77.8&52.3\\
    LAION-B/32\cite{schuhmann2022laion}& & &86.2&63.4&60.2&52.3&70.8&46.5\\
    \midrule
    OpenAI-B/16\cite{radford2021learning}& & &86.2&63.4&68.3&61.9&77.7&48.2\\
    OpenAI-B/32\cite{radford2021learning}&WIT-400M&13&86.2&63.4&63.3&55.9&69.3&42.3\\
    OpenAI-RN50\cite{radford2021learning}& & &38.3&63.4&59.8&52.8&60.7&35.4\\
    
    \midrule
    \midrule
    OpenCLIP-RN50\cite{ilharco_gabriel_2021_5143773_openclip}& CC12M&(-) &38.3&63.4&35.9&30.1&44.7&23.5\\
    \midrule
    OpenCLIP-RN101\cite{ilharco_gabriel_2021_5143773_openclip}& YFCC15M &(-) &56.3&63.4&34.1&29.3&26.3&8.9\\
    \midrule
    CLIPKD-B/16\cite{yang2024clip} & & &86.2&37.8& 37.0&32.1&48.4&26.0\\
    CLIPKD-RN101\cite{yang2024clip}& & &56.3&37.8& 36.8&31.9&49.2&26.7\\
    CLIPKD-Swin/T\cite{yang2024clip}& CC12M+CC3M &0.48&27.9&21.3& 40.2&34.9&51.4&28.2\\
    CLIPKD-MobileNetV3\cite{yang2024clip}& & &2.0&21.3& 27.0&23.0&30.6&14.1\\
    CLIPKD-RN18\cite{yang2024clip}& & &11.4&21.3&31.4&26.9&39.2&20.0\\
    \midrule
    \textbf{SiCLIP(ours)}&CC12M-SYN&\textbf{0.38}& \underline{9.78}&\underline{42.4}&\textbf{\underline{58.1}}&\textbf{\underline{50.7}}&\textbf{\underline{72.5}}&\textbf{\underline{47.3}}\\
    
    \bottomrule
  \end{tabular}
    \caption{Zero-shot classification performance compared with existing works.}
      \label{tab:classification}
\end{table*}

\section{Experiments}

\subsection{Implementation Details}
We adopt a warmup strategy during the first 10000 training iterations. We use AdamW optimizer, and set the batch size to 1536 and weight-decay to 0.1. We train our model for 32 epochs with learning rate 0.001 on an Nvidia RTX3090. In ablation studies, the number of epochs is set to 9. We adopt MobileCLIP-S0 as the teacher for WIKD. For hyper-parameters, we set $\lambda_1=4000$, $\lambda_2=\lambda_3=1$, $\lambda_4=0.1$. Other settings follow CLIP-KD \cite{yang2024clip}.

We evaluate the zero-shot performance on multiple datasets. Specifically, we use ImageNet-1k \cite{deng2009imagenet}, ImageNet-V2 \cite{recht2019imagenetv2}, ImageNet-R \cite{hendrycks2021manyimagenetr} and ImageNet-S \cite{wang2019learningimagenets} to evaluate the zero-shot image classification performance. And for zero-shot image-text retrieval, we use MSCOCO \cite{lin2014microsoftcoco} and Flickr30k \cite{plummer2015flickr30k}. By default, we report top-1 accuracy (acc1) for image classification and R@1 for image-text retrieval.

\subsubsection{Data Augmentation.}
We apply RandomResizedCrop and RandAugment \cite{cubuk2020randaugment} for image augmentation. We set the scale to (0.08, 1.0) in RandomResizedCrop for a strong augmentation on the original image, and then apply RandAugment on the processed image, which further augments the image by randomly adopting two of the default 31 augmentation methods \cite{cubuk2020randaugment}


\subsection{Main Results}
\subsubsection{Zero-shot image-text retrieval.}
Table \ref{tab:retrieval} reports the zero-shot image-text retrieval performance on MSCOCO and Flickr30k. Compared with models trained on similar-scale datasets (up to 20M samples), our model outperforms all other works on MSCOCO. As for Flickr30k, our model also achieves the performance of current state-of-the-art model TinyCLIP while using fewer parameters. 
Compared with models trained on large-scale datasets (400M-1B), our model achieves competitive performance and outperforms many existing works. For example, compared to state-of-the-art MobileCLIP-S0,  our model achieves approximately only 1\% lower T2I performance while using approximately 3\% training samples and 14\% fewer image encoder parameters. Moreover, our model outperforms DataComp-B/32, OpenAI-X(except B/16), and LAION-B/32 on both I2T and T2I metrics on both datasets.

\subsubsection{Zero-shot image classification on ImageNet.}
Table \ref{tab:classification} reports the zero-shot classification performances. When compared with models trained on similar-scale datasets, our model outperforms other works on all reported datasets, demonstrating the effectiveness of our methods. As for large-scale datasets, although not as good as the latest state-of-the-art DataComp-B/16, we still achieve some competitive results compared with several existing works.

\subsubsection{Inference Speed.}

\begin{table}[t]
\centering
\setlength{\tabcolsep}{2pt} 
\renewcommand{\arraystretch}{1.2} 
\footnotesize 

\begin{tabular}{|c|cccc|}
\hline
Model&SiCLIP(ours)& Mobile-S0&Mobile-S1&Mobile-S2\\ \hline
Imgs/sec&\textbf{39.5}&38.2(-1.3)&22.2(-17.3)&18.2(-21.3)\\ \hline
\end{tabular}
\caption{Comparison on inference speed.}
\label{tab:inference_speed}
\end{table}

\begin{table}[htbp]
\centering

\setlength{\tabcolsep}{1pt} 
\renewcommand{\arraystretch}{1.2} 

\begin{tabular}{|ccccc|}
\hline
Dataset
(Storage)& IN-1k& \multicolumn{2}{c}{Flickr30k} & \\ 
&&(T$\xrightarrow{}$I)&(I$\xrightarrow{}$T) & \\ \hline
CC12M (235G)    & 53.4 & 57.4 & 73.5 & \\ \hline
CC12M-SYN (236G) & \textbf{54.5}(+1.1) & \textbf{65.1}(+7.7) & \textbf{78.0}(+4.5)&\\ \hline
\end{tabular}
\caption{CC12M-SYN vs CC12M.}
\label{tab:syn_vs_cc12m}
\end{table}
\begin{figure}[htbp]
    \centering
    \includegraphics[width=0.8\linewidth]{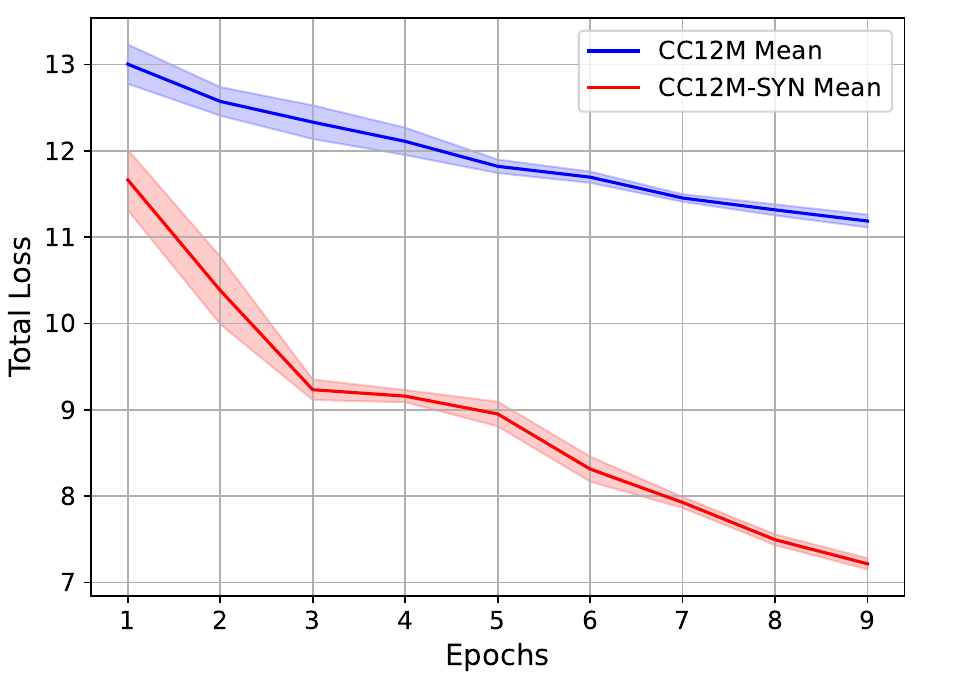}
    \caption{Comparison of training efficiency between CC12M and CC12M-SYN dataset.}
    \label{fig:training_efficiency}
\end{figure}
To evaluate the inference speed, we perform simulation experiments on CPU (Intel(R)-Xeon(R)-Silver-4314-CPU@2.40GHz) and compare the average inference speed with the state-of-the-art MobileCLIP series \cite{vasu2024mobileclip}. As shown in table \ref{tab:inference_speed}, given an input sequence of 1000 images, the processing speed of our model achieves 39.5 images/sec, slightly outperforming state-of-the-art MobileCLIP-S0 (38.2 images/sec). This demonstrates the benefits of adopting SAS-P blocks.

\subsection{Ablation Studies}

\subsubsection{Training Efficiency of CC12M-SYN.}

To demonstrate the training efficiency improvement of CC12M-SYN, we train our model separately on CC12M-SYN and CC12M for 20 epochs. We report the average loss curve of first 9 epochs and zero-shot performance on IN-1k and Flickr30k at the last epoch. Fig. \ref{fig:training_efficiency} reports the loss curves of CC12M and CC12M-SYN, showing that training on CC12M-SYN leads to faster reduction in loss. Table \ref{tab:syn_vs_cc12m} shows that the model trained on CC12M-SYN has better performance on both zero-shot classification and zero-shot image-text retrieval. These results indicate the benefits of synthetic labels for data diversity and quality improvement.

\begin{table}[ht]
\centering

\setlength{\tabcolsep}{3pt} 
\renewcommand{\arraystretch}{1.2} 

\begin{tabular}{|ccccc|}
\hline
Method & IN-1k& \multicolumn{2}{c}{Flickr30k}& \\ 
&&(T$\xrightarrow{}$I)&(I$\xrightarrow{}$T)&\\ \hline
 Baseline&\underline{27.0}&\underline{49.1}&\underline{60.9}& \\ \hline
  WI&40.0(+13.0)&55.2(+6.1)&61.0(+0.1)&\\ \hline
  WIKD & 52.4(+25.4) & 62.8(+13.7) & 76.1(+15.2)&\\ \hline
  WIKD+PM&\textbf{55.0}(+28.0)&\textbf{64.7}(+15.6)&\textbf{79.1}(+18.2)& \\ \hline
\end{tabular}
\caption{Analysis of WIKD and PM loss.}
\label{tab:analysis_of_wikd_pm}
\end{table}
\subsubsection{Analysis of WIKD and PM loss.}
We explore the effectiveness of WIKD and PM loss by comparing the performance of training without WIKD and PM loss (Baseline), training with WI-only, training with WIKD, and training with both WIKD and PM loss. Results are reported in Table \ref{tab:analysis_of_wikd_pm}. It shows that training with WI-only can provide an improvement on both zero-shot classfication and image-text retrieval (+13.0 and +6.1/+0.1 on acc1 of classification and R@1 of retrieval). And when trained with WIKD, the performances will be higher (+25.4 and +15.9/+15.2, respectively). When trained with WIKD and PM loss simultaneously, the model achieves the highest performances. These results evidently support the effectiveness of both WIKD and PM loss.

\section{Conclusion}
In this work, we have proposed a number of techniques to enable training and inference of CLIP models on consumer-level computers while achieving competitive performance. This is critical in bringing the impressive results of foundation models to edge devices.  We have reduced the model structure, improving the inference speed. Moreover, we proposed WIKD and PM loss, which have contributed to performance improvement, and can be used in simplifying other models in different domains. Finally, trained on the augmented CC12M-SYN dataset, our model achieved competitive performance compared to existing works despite having fewer parameters and being trained on a smaller dataset.

\bigskip
\bibliographystyle{unsrt}
\bibliography{ref}
\end{document}